\documentclass[manuscript]{acmart}

\usepackage[ruled,vlined]{algorithm2e}   
\usepackage{subfig}
\AtBeginDocument{%
  \providecommand\BibTeX{{%
    \normalfont B\kern-0.5em{\scshape i\kern-0.25em b}\kern-0.8em\TeX}}}

\setcopyright{acmcopyright}

\copyrightyear{2021} \acmYear{2021} \setcopyright{rightsretained}

\acmDOI{10.1145/1122445.1122456}
\acmConference[IUI '21 Companion]{26th International Conference on Intelligent User Interfaces}{April 14--17, 2021}{College Station, TX, USA}
\acmBooktitle{26th International Conference on Intelligent User Interfaces(IUI '21 Companion), April 14--17, 2021, College Station, TX, USA}
\acmDOI{10.1145/3397482.3450717}
\acmISBN{978-1-4503-8018-8/21/04}

\begin{document}

\title{User Preferential Tour Recommendation Based on POI-Embedding Methods}

\author{Ngai~Lam~Ho}
\email{ngailam\_ho@mymail.sutd.edu.sg}
\orcid{0000-0003-4768-2208}
\affiliation{%
  \department{Information Systems Technology and Design Pillar}
  \institution{Singapore University of Technology and Design}
  \country{Singapore}
}

\author{Kwan Hui Lim}
\email{kwanhui\_lim@sutd.edu.sg}
\orcid{0000-0002-4569-0901}
\affiliation{%
  \department{Information Systems Technology and Design Pillar}
  \institution{Singapore University of Technology and Design}
  \country{Singapore}
}

\renewcommand{\shortauthors}{Ho and Lim}

\begin{abstract}
Tour itinerary planning and recommendation are challenging tasks for tourists in unfamiliar countries.
 Many tour recommenders only consider broad POI categories and do not align well with users’ 
 preferences and other locational constraints. We propose an algorithm to recommend personalized tours using POI-embedding methods, which provides a finer representation of POI types. Our recommendation algorithm will generate a sequence of POIs that optimizes time and locational constraints, as well as user’s preferences based on past trajectories from similar tourists. Our tour recommendation algorithm is modelled as a word embedding model in natural language processing, coupled with an iterative algorithm for generating itineraries that satisfies time constraints. Using a Flickr dataset of~4~cities, preliminary experimental results show that our algorithm is able to recommend a relevant and accurate itinerary, based on measures of recall, precision and \textsl{F1}-scores.

\end{abstract}



\maketitle

\section{Introduction} 
Tour recommendation is an important task for tourists to visit unfamiliar places~\cite{he2017category,lim2019tour}. Tour recommendation and planning are challenging problems due to time and locality constraints faced by the tourists visiting unfamiliar cities~\cite{brilhante-ipm15,chen-cikm16,gionis-wsdm14}. 
Most visitors usually follow guide books/websites to plan their daily itineraries or use recommendation systems that suggest places-of-interest (POIs) based on popularity~\cite{lim2019tour}. However, these are not optimized in terms of time feasibility, localities and users’ preferences. Tourists visiting an unfamiliar city are usually constrained by time, such as hotel bookings or air flight itineraries.
In this paper, we propose a word embedding model to recommend POIs based on historical data and their popularity with consideration of the locations and traveling time between these POIs. We combine tour recommendation with various word-embedding model, namely Skip-Gram~\cite{mikolov2013distributed}, Continuous ~Bag~of~Words~\cite{mikolov2013efficient} and \textsl{FastText}~\cite{bojanowski2017enriching}, in the tour recommendation problem. The results show our algorithm can achieve \textsl{F1}-scores of up to 59.2\% accuracy in our experiments. 

\section{Related Work} 
Tour planning is an essential task for tourists.
Most visitors rely on guide books or websites to plan their daily itineraries which can be time-consuming. Next POI prediction~\cite{he2017category,zhao2020go} and tour planning~\cite{sohrabi2020greedy,lim2019tour} are two related problems: Next-location prediction and recommendation aim to identify the next POI that is most likely to visit based on historical trajectories. Tour itinerary recommendation aims to recommend multiple POIs or locations in the form of a trajectory. On the other hand, Top-$k$ location recommendation provides recommendation to multiple POIs, but they do not provide a solution to these POIs as a connected itinerary. Furthermore, tour itinerary recommendation has the additional challenges of planning an itinerary of connected POIs that appeal to the interest preferences of the users, satisfying tourists' temporal and spatial constraints in the form of a limited time budget. Various works have utilized geotag photos to determine POI related information for making various types of tour recommendation~\cite{lim2018personalized,cai2018itinerary,kurashima2013travel,sun2017tour}.


\section{Problem Formulation and Algorithm} 

\textbf{Formulation~}{
  We denote a traveler, $u$, visiting $k$ POIs in a city, in a sequence of $(poi,time)$~tuples,~$S_u = [ (p_1,t_1),(p_2,t_2)...$ $(p_k,t_k)]$, where $k=|S_u|$,~for all~$p_i \in POIs$ and ${t_i}$ as the timestamps of the photos taken.  Given also, a starting POI-${s_0} \in {POIs}$, the problem in this paper is to  recommend a sequence of POIs which travelers are more \emph{likely} to visit using word embedding methods.
} \\
\textbf{Algorithm~~~~}{
Our algorithm is adapted from the \emph{word2vec}~model where we treat POIs to be analogous to words in its typical NLP application, i.e., POIs are akin to words, and itineraries to sentences. To measure POI-POI similarity, we first convert travel trajectories to a \textit{word2vec}~model, by analyzing the past  activities of $n$~users moving from  $p_i$ to $p_{i+1}$ by considering first event that is at least eight hours from his/her previous activity~(i.e. minimum rest time of 8 hours.)  We next construct a set of \emph{sentences} of POIs in our embedding model, as an input to the $word2vec$~model~(also known as \emph{corpus}), as shown in Algorithm~1.   The  model is then trained using different \emph{word2vec}/\textsl{FastText}~(FT) models and different hyper-parameters~(such as dimensionalities/~Epoches) to describe their POI-POI similarities in our travel recommendation model. We then outline the prediction algorithm using \emph{word2vec} embedding models, with some initial starting location~($POI_1$).

\begin{figure*}[h]
    \centering
       \includegraphics[
     trim=2mm 3mm 83mm 18mm, clip, width=0.49\textwidth, clip=true] {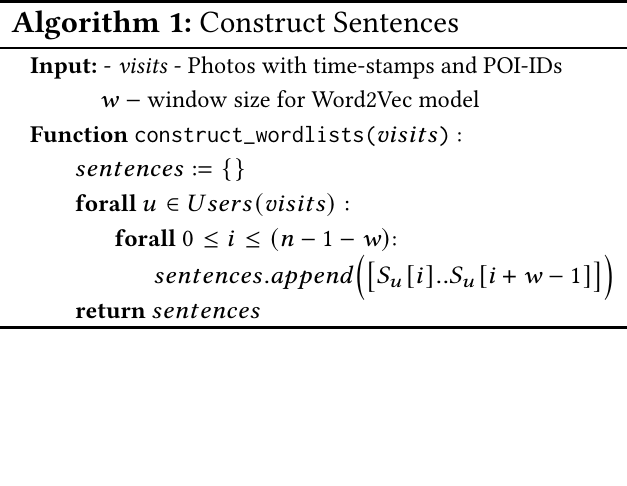}
       \includegraphics[
     trim=78mm 2mm 5mm 1mm, clip, width=0.49\textwidth, clip=true] {algorithms_1-2}
\end{figure*}

Algorithm~2 recommends popular POIs based on previously visited POIs and present location by \emph{iteratively} recommending the \emph{closest} POI in terms of \textsl{cosine} similarities, to its present location, but \emph{farthest} from the past POIs.
  We also evaluated $FastText$ which uses \emph{character-based} $n$-grams for measuring similarity  between POI-vectors. Since $FastText$ considers sub-words (i.e.~partial POI~names) in its embedding model, using Skip-Gram or CBOW at character level, it can capture the meanings of suffixes/prefixes in POI names more accurately. Moreover, it is also useful in handling situations for POIs that are not found in the  \emph{corpus}.


\section{Experiments and Results}

\subsection{Datasets and Baseline Algorithm}
We use the dataset from~\cite{lim2018personalized} comprising the travel histories of 5,654 users from Flickr in~4 different cities with meta information, such as the date/time and geo-location coordinates.
Using this dataset, we constructed the travel histories by chronologically sorting the photos, which resulted in the users' trajectories. These trajectories are then regarded as sentences as inputs to our POI-embedding models for training.}
As a baseline algorithm for comparison, we use a greedy heuristic algorithm that commences a trip from a starting POI $p_1$ and iteratively choose to visit an \emph{un-visited} POI with the most number of photos posted~\cite{Liu-ECMLPKDD20}. The sequence of selected POIs forms the recommended itinerary based on the \emph{popularities} of the POIs. In our experiment, we used daily itineraries from users; by considering the start of a day tour that is at least 8 hours from the last photo posted.


\begin{figure*}[h]
\label{fig:route}
    \centering
    \subfloat{{ 
       \includegraphics[
       trim=135mm 82mm 60mm 87mm, clip, width=0.47\textwidth, clip=true] {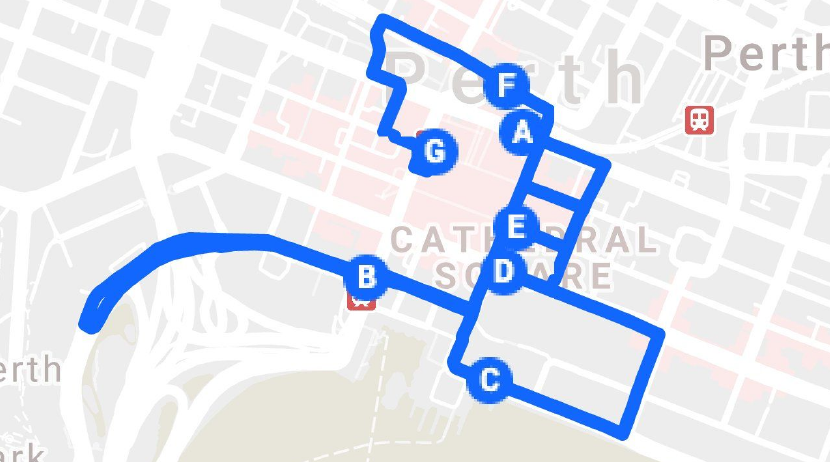}}}
    \qquad 
    \subfloat{{
       \includegraphics[
       trim=135mm 82mm 60mm 87mm, clip, width=0.47\textwidth, clip=true] {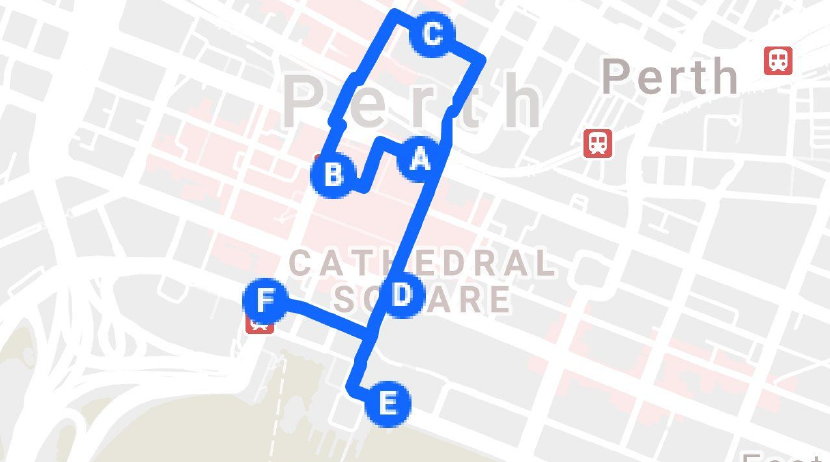}}}
    \label{fig:example}%
 \par
 \caption{
    A user’s actual travel trajectory~(left) vs. Route recommended by the proposed algorithm~(right) in Perth.
    \small
  \emph{User's route:} 
     Wellington station~$\rhd $ 
     Queens gardens~$\rhd$ 
     Supreme court gardens~$\rhd$ 
     Stirling gardens~$\rhd$ 
     Perth town hall~$\rhd$ 
     West Aust. museum~$\rhd$ 
     Railway station~/
  \emph{Recommended route:}
    Wellington station~$\rhd$ 
    West Aust. museum~$\rhd$ 
    State Library~$\rhd$
    Stirling gardens~$\rhd$ 
    Supreme court gardens~$\rhd$ 
    Esplanade station.
  }
\end{figure*}

\subsection{Experiments and Results}
  We describe the algorithm of getting the context words to train our \textit{word2vec}~models in Algorithm~1.  These travel histories regarded as sequences of POI names are trained with \textit{word2vec}~models, i.e. \textsl{Skip-Gram}, \textsl{CBOW} and \textsl{FastText} by treating each POI-name composed of character-level \emph{n-}grams. 
  We evaluate the effectiveness of our prediction algorithm in terms of \textsl{precision}~($T_P$), \textsl{recall}~($T_R$) and \textsl{F1-scores} of our predicted POIs again the actual POIs in the actual travel sequences. Let $S_p$ be the predicted sequence of POIs from the algorithm and $S_u$ be the actual sequence from the users, we evaluate our algorithms based on: 
  $T_R(S_u,S_p)$ = $ \frac{|S_u \cap S_p|}
                              {|S_p|}$,
  $T_P(S_u,S_p) = \frac{|S_u \cap S_p|}{|S_u|}$, and 
  $F1\_score(S_u,S_p) =  \frac{2  T_R(\bullet)  T_P(\bullet)}
                              {T_R(\bullet) + T_P(\bullet)}$.
  We summarize the results of our method using different \textit{word2vec} models and hyper-parameters by varying epoch, windows~size and dimensionality our data datasets.
  As a use case of our recommended tour itinerary, Fig.1 shows a route recommended by our algorithm~(right) in the city of Perth, while one of the users' itinerary is plotted on the left. In addition to recommending an itinerary with more relevant POIs, our recommendation results in an itinerary that is more compact in terms of travelling path, which translates to less time  spent travelling for the tourist. Figure~2 shows a more detailed break-down of our experimental results in terms of  recall, precision and F1 scores for the 4 cities. Our proposed algorithm can optimally suggest POIs based on users' present and past locations against historical dataset.

\begin{figure*}[h]
  \label{fig:results}
    \includegraphics[
     trim=5 5 5 1, clip, width=\textwidth, clip=true]
     {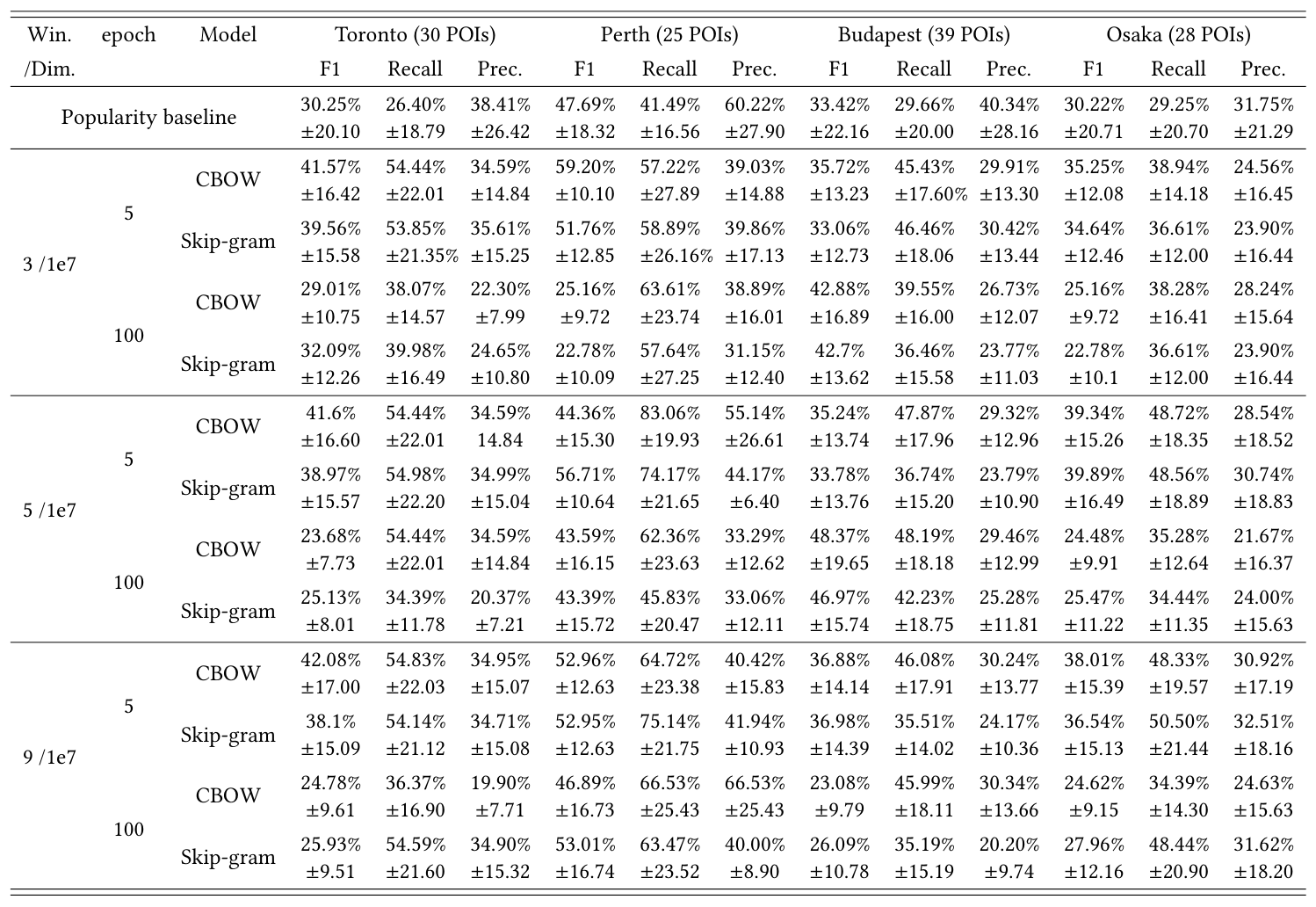}
    \caption{Average F1/Recall/Precision-scores of embedding methods \
       {\small       
          (with different hyper-parameters) and baseline algorithm in 4~cities  (Win~:~window size of \textit{word2vec} in sentence.    Dim.~:~maximum dimensionality of the \textit{word2vec} vector).  }  }
\end{figure*} \vspace{-10px}

\section{Conclusion}
In this paper, we study the problem of tour itinerary recommendation. We propose an algorithm that translates travel trajectories into word-vector space, followed by an iterative heuristic algorithm that constructs itineraries constrained by time and space. Our prediction algorithm reliably uncover user preference in a tour by using one POI of users' preference.  Our preliminary experiments show efficient and reliable methods for predicting popular  POIs in terms of precision, recall, and \textsl{F1}-scores. In our experiments on 4~cities, our proposal POI-embedding algorithm  outperforms baseline algorithm based on POI~popularity. Future work would include further evaluation of scoring the prediction algorithms and experiment on more cities and POI sets. In future, we will extend the algorithms to incorporate more dimensions into the \textit{word2vec} models such as spatiality and popularity.

\begin{acks}
This research is funded in part by the Singapore University of Technology and Design under grant SRG-ISTD-2018-140.
The computational work was partially performed on resources of the National Supercomputing Centre, Singapore.
\end{acks}

\bibliographystyle{ACM-Reference-Format}
\bibliography{ref}


\begin{thebibliography}{15}


\ifx \showCODEN    \undefined \def \showCODEN     #1{\unskip}     \fi
\ifx \showDOI      \undefined \def \showDOI       #1{#1}\fi
\ifx \showISBNx    \undefined \def \showISBNx     #1{\unskip}     \fi
\ifx \showISBNxiii \undefined \def \showISBNxiii  #1{\unskip}     \fi
\ifx \showISSN     \undefined \def \showISSN      #1{\unskip}     \fi
\ifx \showLCCN     \undefined \def \showLCCN      #1{\unskip}     \fi
\ifx \shownote     \undefined \def \shownote      #1{#1}          \fi
\ifx \showarticletitle \undefined \def \showarticletitle #1{#1}   \fi
\ifx \showURL      \undefined \def \showURL       {\relax}        \fi
\providecommand\bibfield[2]{#2}
\providecommand\bibinfo[2]{#2}
\providecommand\natexlab[1]{#1}
\providecommand\showeprint[2][]{arXiv:#2}

\bibitem[\protect\citeauthoryear{Bojanowski, Grave, Joulin, and
  Mikolov}{Bojanowski et~al\mbox{.}}{2017}]%
        {bojanowski2017enriching}
\bibfield{author}{\bibinfo{person}{Piotr Bojanowski}, \bibinfo{person}{Edouard
  Grave}, \bibinfo{person}{Armand Joulin}, {and} \bibinfo{person}{Tomas
  Mikolov}.} \bibinfo{year}{2017}\natexlab{}.
\newblock \showarticletitle{Enriching word vectors with subword information}.
\newblock \bibinfo{journal}{\emph{Transactions of the Association for
  Computational Linguistics}}  \bibinfo{volume}{5} (\bibinfo{year}{2017}),
  \bibinfo{pages}{135--146}.
\newblock


\bibitem[\protect\citeauthoryear{Brilhante, Macedo, Nardini, Perego, and
  Renso}{Brilhante et~al\mbox{.}}{2015}]%
        {brilhante-ipm15}
\bibfield{author}{\bibinfo{person}{Igo~Ramalho Brilhante},
  \bibinfo{person}{Jose~Antonio Macedo}, \bibinfo{person}{Franco~Maria
  Nardini}, \bibinfo{person}{Raffaele Perego}, {and} \bibinfo{person}{Chiara
  Renso}.} \bibinfo{year}{2015}\natexlab{}.
\newblock \showarticletitle{On planning sightseeing tours with
  {T}rip{B}uilder}.
\newblock \bibinfo{journal}{\emph{Information Processing \& Management}}
  \bibinfo{volume}{51}, \bibinfo{number}{2} (\bibinfo{year}{2015}),
  \bibinfo{pages}{1--15}.
\newblock


\bibitem[\protect\citeauthoryear{Cai, Lee, and Lee}{Cai et~al\mbox{.}}{2018}]%
        {cai2018itinerary}
\bibfield{author}{\bibinfo{person}{Guochen Cai}, \bibinfo{person}{Kyungmi Lee},
  {and} \bibinfo{person}{Ickjai Lee}.} \bibinfo{year}{2018}\natexlab{}.
\newblock \showarticletitle{Itinerary recommender system with semantic
  trajectory pattern mining from geo-tagged photos}.
\newblock \bibinfo{journal}{\emph{Expert Systems with Applications}}
  \bibinfo{volume}{94} (\bibinfo{year}{2018}), \bibinfo{pages}{32--40}.
\newblock


\bibitem[\protect\citeauthoryear{Chen, Ong, and Xie}{Chen
  et~al\mbox{.}}{2016}]%
        {chen-cikm16}
\bibfield{author}{\bibinfo{person}{Dawei Chen}, \bibinfo{person}{Cheng~Soon
  Ong}, {and} \bibinfo{person}{Lexing Xie}.} \bibinfo{year}{2016}\natexlab{}.
\newblock \showarticletitle{Learning Points and Routes to Recommend
  Trajectories}. In \bibinfo{booktitle}{\emph{Proceedings of the 25th ACM
  CIKM'16}}. \bibinfo{pages}{2227--2232}.
\newblock


\bibitem[\protect\citeauthoryear{Gionis, Lappas, Pelechrinis, and Terzi}{Gionis
  et~al\mbox{.}}{2014}]%
        {gionis-wsdm14}
\bibfield{author}{\bibinfo{person}{Aristides Gionis},
  \bibinfo{person}{Theodoros Lappas}, \bibinfo{person}{Konstantinos
  Pelechrinis}, {and} \bibinfo{person}{Evimaria Terzi}.}
  \bibinfo{year}{2014}\natexlab{}.
\newblock \showarticletitle{Customized tour recommendations in urban areas}. In
  \bibinfo{booktitle}{\emph{Proceedings of WSDM'14}}.
  \bibinfo{pages}{313--322}.
\newblock


\bibitem[\protect\citeauthoryear{He, Li, and Liao}{He et~al\mbox{.}}{2017}]%
        {he2017category}
\bibfield{author}{\bibinfo{person}{Jing He}, \bibinfo{person}{Xin Li}, {and}
  \bibinfo{person}{Lejian Liao}.} \bibinfo{year}{2017}\natexlab{}.
\newblock \showarticletitle{Category-aware next point-of-interest
  recommendation via listwise Bayesian personalized ranking}. In
  \bibinfo{booktitle}{\emph{Proceedings of the 26th International Joint
  Conference on Artificial Intelligence}}. \bibinfo{pages}{1837--1843}.
\newblock


\bibitem[\protect\citeauthoryear{Kurashima, Iwata, Irie, and
  Fujimura}{Kurashima et~al\mbox{.}}{2013}]%
        {kurashima2013travel}
\bibfield{author}{\bibinfo{person}{Takeshi Kurashima},
  \bibinfo{person}{Tomoharu Iwata}, \bibinfo{person}{Go Irie}, {and}
  \bibinfo{person}{Ko Fujimura}.} \bibinfo{year}{2013}\natexlab{}.
\newblock \showarticletitle{Travel route recommendation using geotagged
  photos}.
\newblock \bibinfo{journal}{\emph{Knowledge and information systems}}
  \bibinfo{volume}{37}, \bibinfo{number}{1} (\bibinfo{year}{2013}),
  \bibinfo{pages}{37--60}.
\newblock


\bibitem[\protect\citeauthoryear{Lim, Chan, Karunasekera, and Leckie}{Lim
  et~al\mbox{.}}{2019}]%
        {lim2019tour}
\bibfield{author}{\bibinfo{person}{Kwan~Hui Lim}, \bibinfo{person}{Jeffrey
  Chan}, \bibinfo{person}{Shanika Karunasekera}, {and}
  \bibinfo{person}{Christopher Leckie}.} \bibinfo{year}{2019}\natexlab{}.
\newblock \showarticletitle{Tour recommendation and trip planning using
  location-based social media: a survey}.
\newblock \bibinfo{journal}{\emph{Knowledge and Information Systems}}
  (\bibinfo{year}{2019}), \bibinfo{pages}{1--29}.
\newblock


\bibitem[\protect\citeauthoryear{Lim, Chan, Leckie, and Karunasekera}{Lim
  et~al\mbox{.}}{2018}]%
        {lim2018personalized}
\bibfield{author}{\bibinfo{person}{Kwan~Hui Lim}, \bibinfo{person}{Jeffrey
  Chan}, \bibinfo{person}{Christopher Leckie}, {and} \bibinfo{person}{Shanika
  Karunasekera}.} \bibinfo{year}{2018}\natexlab{}.
\newblock \showarticletitle{Personalized trip recommendation for tourists based
  on user interests, points of interest visit durations and visit recency}.
\newblock \bibinfo{journal}{\emph{Knowledge and Information Systems}}
  \bibinfo{volume}{54}, \bibinfo{number}{2} (\bibinfo{year}{2018}),
  \bibinfo{pages}{375--406}.
\newblock


\bibitem[\protect\citeauthoryear{Liu, Wood, and Lim}{Liu et~al\mbox{.}}{2020}]%
        {Liu-ECMLPKDD20}
\bibfield{author}{\bibinfo{person}{Junhua Liu}, \bibinfo{person}{Kristin~L.
  Wood}, {and} \bibinfo{person}{Kwan~Hui Lim}.}
  \bibinfo{year}{2020}\natexlab{}.
\newblock \showarticletitle{{Strategic and Crowd-Aware Itinerary
  Recommendation}}. In \bibinfo{booktitle}{\emph{{Proceedings of the 2020
  ECML-PKDD'20}}}.
\newblock


\bibitem[\protect\citeauthoryear{Mikolov, Chen, Corrado, and Dean}{Mikolov
  et~al\mbox{.}}{2013a}]%
        {mikolov2013efficient}
\bibfield{author}{\bibinfo{person}{Tomas Mikolov}, \bibinfo{person}{Kai Chen},
  \bibinfo{person}{Greg Corrado}, {and} \bibinfo{person}{Jeffrey Dean}.}
  \bibinfo{year}{2013}\natexlab{a}.
\newblock \showarticletitle{Efficient Estimation of Word Representations in
  Vector Space}. In \bibinfo{booktitle}{\emph{Proceedings of the International
  Conf. on Learning Representations}}.
\newblock


\bibitem[\protect\citeauthoryear{Mikolov, Sutskever, Chen, Corrado, and
  Dean}{Mikolov et~al\mbox{.}}{2013b}]%
        {mikolov2013distributed}
\bibfield{author}{\bibinfo{person}{Tomas Mikolov}, \bibinfo{person}{Ilya
  Sutskever}, \bibinfo{person}{Kai Chen}, \bibinfo{person}{Greg~S Corrado},
  {and} \bibinfo{person}{Jeff Dean}.} \bibinfo{year}{2013}\natexlab{b}.
\newblock \showarticletitle{Distributed representations of words and phrases
  and their compositionality}. In \bibinfo{booktitle}{\emph{Advances in neural
  information processing systems}}. \bibinfo{pages}{3111--3119}.
\newblock


\bibitem[\protect\citeauthoryear{Sohrabi, Ziarati, and Keshtkaran}{Sohrabi
  et~al\mbox{.}}{2020}]%
        {sohrabi2020greedy}
\bibfield{author}{\bibinfo{person}{Somayeh Sohrabi}, \bibinfo{person}{Koorush
  Ziarati}, {and} \bibinfo{person}{Morteza Keshtkaran}.}
  \bibinfo{year}{2020}\natexlab{}.
\newblock \showarticletitle{A Greedy Randomized Adaptive Search Procedure for
  the Orienteering Problem with Hotel Selection}.
\newblock \bibinfo{journal}{\emph{European Journal of Operational Research}}
  \bibinfo{volume}{283}, \bibinfo{number}{2} (\bibinfo{year}{2020}),
  \bibinfo{pages}{426--440}.
\newblock


\bibitem[\protect\citeauthoryear{Sun and Lee}{Sun and Lee}{2017}]%
        {sun2017tour}
\bibfield{author}{\bibinfo{person}{Chih-Yuan Sun} {and}
  \bibinfo{person}{Anthony~JT Lee}.} \bibinfo{year}{2017}\natexlab{}.
\newblock \showarticletitle{Tour recommendations by mining photo sharing social
  media}.
\newblock \bibinfo{journal}{\emph{Decision Support Systems}}
  \bibinfo{volume}{101} (\bibinfo{year}{2017}).
\newblock


\bibitem[\protect\citeauthoryear{Zhao, Luo, Liu, Zhuang, Xu, Li, Sheng, and
  Zhou}{Zhao et~al\mbox{.}}{2020}]%
        {zhao2020go}
\bibfield{author}{\bibinfo{person}{Pengpeng Zhao}, \bibinfo{person}{Anjing
  Luo}, \bibinfo{person}{Yanchi Liu}, \bibinfo{person}{Fuzhen Zhuang},
  \bibinfo{person}{Jiajie Xu}, \bibinfo{person}{Zhixu Li},
  \bibinfo{person}{Victor~S Sheng}, {and} \bibinfo{person}{Xiaofang Zhou}.}
  \bibinfo{year}{2020}\natexlab{}.
\newblock \showarticletitle{Where to go next: A spatio-temporal gated network
  for next poi recommendation}.
\newblock \bibinfo{journal}{\emph{IEEE TKDE}} (\bibinfo{year}{2020}).
\newblock


\end{thebibliography}

\end{document}